\relax
\documentclass[letterpaper,twocolumn]{article} 
\usepackage{simpleConference}  
\usepackage{times}  
\usepackage{helvet}  
\usepackage{courier}  
\usepackage[hyphens]{url}  
\usepackage{graphicx} 
\usepackage[numbers]{natbib}  
\usepackage{caption} 
\DeclareCaptionStyle{ruled}{labelfont=normalfont,labelsep=colon,strut=off} 
\usepackage{algorithm}
\usepackage{algorithmic}

\usepackage{newfloat}
\usepackage{listings}
\floatstyle{ruled}
\newfloat{listing}{tb}{lst}{}
\floatname{listing}{Listing}

\usepackage{amsmath}
\usepackage{amssymb}
\usepackage{amsthm}
\usepackage{hyperref}
\hypersetup{
    colorlinks=true,
    urlcolor=cyan,
}

\graphicspath{ {./} }
\newtheorem{theorem}{Theorem}

\title{SLAW: Scaled Loss Approximate Weighting for Efficient Multi-Task Learning}
\author{
    Michael Crawshaw, Jana Ko\v{s}eck\'a \\
    \\
    George Mason University \\
    mcrawshaw@gmu.edu, kosecka@cs.gmu.edu
}

\begin{document}

\maketitle

\begin{abstract}
Multi-task learning (MTL) is a subfield of machine learning with important applications,
but the multi-objective nature of optimization in MTL leads to difficulties in balancing
training between tasks. The best MTL optimization methods require individually computing
the gradient of each task's loss function, which impedes scalability to a large number
of tasks. In this paper, we propose Scaled Loss Approximate Weighting (SLAW), a method
for multi-task optimization that matches the performance of the best existing methods
while being much more efficient. SLAW balances learning between tasks by estimating the
magnitudes of each task's gradient without performing any extra backward passes. We
provide theoretical and empirical justification for SLAW's estimation of gradient
magnitudes. Experimental results on non-linear regression, multi-task computer vision,
and virtual screening for drug discovery demonstrate that SLAW is significantly more
efficient than strong baselines without sacrificing performance and applicable to a
diverse range of domains.
\end{abstract}

\section{Introduction} \label{introduction}
As machine learning models become more powerful, they gain the ability to perform tasks
which were previously thought impossible. With a growing pool of potential tasks, it is
less feasible to obtain a labeled dataset and train a model from scratch for each new
task. Multi-task learning (MTL) \cite{caruana_1997} can alleviate this scalability issue
by training models to perform multiple tasks simultaneously, sharing parameters and
supervision. Along with increased data and compute efficiency, knowledge sharing can
create an implicit regularization which leads to performance improvements over
single-task models \cite{standley_2019}.

Unfortunately, achieving this boost in efficiency and performance through MTL isn't
easy, due to the fact that multi-task optimization requires optimizing several
(potentially competing) objectives. Previous methods have been proposed to ease the
difficulties of multi-task optimization by balancing learning between tasks
\cite{kendall_2017, chen_2017, liu_2019a, zheng_2018} or addressing negative transfer
\cite{yu_2020a, liu_2020}, where learning on one task hinders learning on another.

The best performing of these recent methods, such as GradNorm \cite{chen_2017}, PCGrad
\cite{yu_2020a}, and IMTL \cite{liu_2020} compute the gradients of each individual
task's loss function, which requires one additional backwards pass per task for each
training step. Even when these methods are sped up by only computing gradients for a
subset of network weights, scalability to a large number of tasks is still an issue,
since the number of required gradient computations scales linearly with the number of
tasks.

To address this issue of scalability, we introduce Scaled Loss Approximate Weighting
(SLAW), an efficient method for multi-task optimization that can be applied in any
gradient-based multi-task learning setting. Instead of computing task-specific
gradients, SLAW balances learning between tasks by assigning weight to tasks based on
the estimated magnitude of each task's gradient. This estimation is theoretically and
empirically justified, and by shedding the need to directly compute gradients of each
task's loss function, SLAW is faster than existing methods and better able to scale to a
large number of tasks. We also show that SLAW is closely related to existing
optimization methods. In particular, we demonstrate that SLAW directly estimates a
solution to a special case of the optimization problem iteratively solved by GradNorm
\cite{chen_2017}.

Our experiments show that SLAW performs as well or better than existing methods while
being much more efficient. For a collection of 10 non-linear regression tasks, SLAW
converges to the ideal optimization trade-off between tasks. For joint semantic
segmentation, surface normal estimation, and depth prediction on NYUv2
\cite{silberman_2012}, SLAW performs as well or better than GradNorm and PCGrad while
running about twice as fast. Finally, for a collection of 128 virtual screening tasks
for drug discovery with the PubChem BioAssay (PCBA) dataset \cite{wang_2016}, SLAW runs
24.4 times faster than GradNorm and 33.0 times faster than PCGrad, while outperforming
GradNorm and being competitive with PCGrad.

Our contributions can be summarized as follows:
\begin{itemize}
    \item We propose Scaled Loss Approximate Weighting (SLAW), an efficient, simple,
        and general method for multi-task optimization. Unlike existing methods, SLAW
        does not compute the gradient of each task's loss function, which greatly
        reduces computational cost.
    \item We provide theoretical and empirical justification for SLAW's estimation of
        the magnitudes of task-specific gradients, demonstrating that SLAW can
        accurately balance learning between tasks without performing additional
        backwards passes.
    \item We analyze the relationship between SLAW and existing methods. In particular,
        we show that SLAW directly estimates a solution to a special case of the
        optimization problem that GradNorm \cite{chen_2017} iteratively solves to
        balance learning between tasks.
    \item We provide experiments on a collection of non-linear regression tasks,
        NYUv2 \cite{silberman_2012}, and a collection of virtual screening tasks from
        PCBA \cite{wang_2016} showing that SLAW quickly learns ideal loss weights, has
        strong overall performance, exhibits efficiency and scalability to a large
        number of tasks, and is applicable to many different domains.
\end{itemize}

The rest of this paper is structured as follows. In Section \ref{related_work}, we
review work related to ours. Section \ref{methods} provides background, a derivation of
SLAW, and an analysis of SLAW's relationship with existing methods. We present
experimental evaluations in Section \ref{experiments}, and conclude with Section
\ref{conclusion}.

\section{Related Work} \label{related_work}
Multi-task learning \cite{caruana_1997} has a long history both within and outside of
deep learning. MTL eases deep learning's need for huge amounts of training data and to
start learning each new task from scratch. Training shared parameters on multiple tasks
allows for supervision of one task to aid in the learning for another, and a set of
trained shared features can often be reused to instantiate learning on a new task,
yielding faster learning through feature reuse \cite{parisotto_2015}. However, achieving
increased data efficiency, transfer robustness, and regularization from MTL is never
guaranteed and highly dependent on the relationships between the tasks involved
\cite{standley_2019, alonso_2016}.

Many families of methods for MTL have arisen, including multi-task architectures
\cite{misra_2016, ruder_2019, pinto_2017, lu_2017}, optimization strategies
\cite{duong_2015, chaudhry_2018, kendall_2017, yu_2020a}, and methods for learning task
relationships \cite{alonso_2016, zamir_2018, standley_2019, achille_2019}. In this
paper, we focus on optimization. Weighting by Uncertainty \cite{kendall_2017} treats
loss weights as learnable parameters that model the uncertainty for each task. GradNorm
\cite{chen_2017} also treats the weights as learnable parameters, but the weights are
optimized separately from network parameters to encourage ideal magnitudes of each task
loss's gradient. Dynamic Weight Averaging \cite{liu_2019a} (or DWA) was introduced as a
computationally efficient alternative to GradNorm that doesn't deal with individual task
gradients. Recently, methods like PCGrad \cite{yu_2020a} and Impartial MTL
\cite{liu_2020} balance learning across tasks by considering the relationships between
directions of gradients of each task's loss function. For a review of MTL as a whole,
see \cite{zhang_2017, crawshaw_2020}.

MTL has been applied in many domains, such as computer vision \cite{misra_2016}, natural
language processing \cite{collobert_2008}, reinforcement learning \cite{pinto_2017},
bioinformatics \cite{ramsundar_2015}, and more. Of these domains, this paper applies MTL
to computer vision and bioinformatics. Computer vision lends itself well to MTL since
many tasks rely on common visual features. Earlier works \cite{zhang_2014, eigen_2015,
dai_2015} focus on training a shared feature extractor jointly on many tasks. Other
architectures leverage task-specific networks with information flow between networks,
such as Cross-Stitch Networks \cite{misra_2016}, Sluice Networks \cite{ruder_2019}, and
NDDR-CNN \cite{gao_2019}. Bioinformatics has historically been less popular as an
application domain for MTL researchers, though there are some examples, most notably
\cite{ramsundar_2015}. \cite{ching_2018} and \cite{miotto_2017} provide surveys of deep
learning for biology and healthcare.

\section{Methods} \label{methods}

\subsection{Background}
In the usual formulation of machine learning, a parameterized model $f_{\theta}$ is
trained to approximate a target function by finding a value of $\theta$ which minimizes
a loss function $L(\theta)$. In contrast, MTL involves training multiple parameterized
models that share parameters on multiple tasks; $f_{1, \theta}, f_{2, \theta}, ...,
f_{n, \theta}$ are jointly trained by minimizing a sum of loss functions $\sum_{i=1}^n
w_i L_i(\theta)$. Here $L_i$ is a loss function for task $i$ and the coefficients $w_i$
are called loss weights, traditionally hand-picked to control the optimization trade-off
between various task losses. Parameter sharing across tasks decreases memory cost, and
multi-task models may outperform their single-task counterparts. However, there is no
guarantee of this perfomance boost.

Although the coefficients $w_i$ in the MTL loss are traditionally hand-picked, these
loss weights can also be chosen algorithmically to ease multi-task optimization. Several
methods have been proposed to automatically choose loss weights based on task
uncertainty \cite{kendall_2017}, learning speed \cite{chen_2017, liu_2019a, zheng_2018,
liu_2019b}, and task performance \cite{guo_2018}.

Previous work \cite{chen_2017} pointed out that loss weighting can serve to balance the
norms of each task loss's gradient. Specifically, let $g_i = \nabla L_i(\theta)$. If the
task loss functions $L_i$ are such that the norms of the task gradients $\|g_i\|$ are
varying in their magnitude, then the tasks with large $\|g_i\|$ will dominate the
training process. In this case, some tasks are ``left behind" and not trained
adequately. To offset this, loss weights can be chosen so that the weighted task
gradient norms $w_i \|g_i\|$ have similar magnitudes, resulting in all tasks being
trained equally.

\subsection{Scaled Loss Approximate Weighting}
In this section, we derive our algorithm for multi-task optimization: Scaled Loss
Approximate Weighting (SLAW). SLAW is motivated by the idea that loss weighting in MTL
can balance the norms of task loss gradients. However, SLAW does so without explicitly
computing the gradient of each task's loss, significantly improving on the scalability
and efficiency of recent methods without sacrificing performance. In the following, we
omit the dependence of $L_i$ and $f_i$ on $\theta$ for convenience of notation.

The goal of SLAW is to choose loss weights $w_i$ so that the weighted gradient norms of
each task loss are equal:
\begin{align} \label{equal_grads}
    w_i \| g_i \| = w_j \| g_j \| ~\forall i, j
\end{align}
If we treat each $w_i$ as unknown and each $\|g_i\|$ as known, then Equation
\ref{equal_grads} yields a homogeneous linear system of $n-1$ equations with $n$
unknowns, so that there are infinitely many solutions of $w_i$. One such solution is:
\begin{align} \label{naive_clw}
    w_i = \frac{1}{\|g_i\|}
\end{align}
Following previous works \cite{chen_2017, liu_2019a}, we also decouple the global
learning rate from the loss weights $w_i$ by enforcing a constraint on the sum of the
loss weights:
\begin{align} \label{weight_sum_constraint}
    \sum_{i=1}^n w_i = n
\end{align}
So that the mean value of $w_i$ is 1. Adding Equation \ref{weight_sum_constraint} to our
system of equations from Equation \ref{equal_grads}, we obtain a homogeneous linear
system of $n$ equations with $n$ unknowns. Combining Equations \ref{naive_clw} and
\ref{weight_sum_constraint}, we can see that the following values of $w_i$ are a valid
solution to this system:
\begin{align} \label{clw}
    w_i = \frac{n}{\|g_i\|} \bigg/ \sum_{j=1}^n \frac{1}{\|g_j\|}
\end{align}
Setting the loss weights in this way at each training step would satisfy the equal
gradient condition in Equation \ref{equal_grads} as well as the temperature condition in
Equation \ref{weight_sum_constraint}, balancing the task gradients as originally
desired. However, this would require computing the gradient $g_i$ of each task's loss
function.

To avoid this computational cost, we can approximate the RHS of Equation \ref{clw} by
approximating the values of each $\|g_i\|$.  Notice that Equation \ref{clw} only depends
on the \textit{magnitude} of each task's gradient, so that an approximation of $\|g_i\|$
is sufficient to approximate Equation \ref{clw}. The following theorem gives a method
for such an approximation:

\begin{theorem} \label{slaw_thm}
Let $f: \mathbb{R}^d \rightarrow \mathbb{R}$ be continuously differentiable and
Lipschitz continuous with Lipschitz constant $K$. Then for any $\mathbf{x}_0 \in
\mathbb{R}^d$ and $\epsilon > 0$, there exists $\delta > 0$ and constants $K_1, K_2$
(which depend only on $n$, $K$, $\delta$, and $\epsilon$) so that:
\begin{align} \label{slaw_thm_conclusion}
    \left| \text{Var}[f(\mathbf{X})] - (K_1 \| \nabla f(\mathbf{x}_0) \|^2 + K_2) \right| < \epsilon
\end{align}
where $\mathbf{X}$ is a random variable which is uniformly distributed over
$\{\mathbf{x}_0 + \mathbf{d} : \|\mathbf{d}\| < \delta \}$.
\end{theorem}

The proof is given in Appendix \ref{slaw_thm_proof}. Put simply, this theorem tells us
that the standard deviation of a ``well-behaved" function in a neighborhood around a
point is nearly proportional to the norm of the gradient at that point. If we assume
that the task loss functions $\{L_i\}_{i=1}^n$ are continuously differentiable and
Lipschitz continuous (a relatively weak assumption), then we can conclude that the norm
of the task gradient $\|g_i\|$ at a point $\theta_0$ is proportional to the standard
deviation of loss values $L_i(\theta)$ as $\theta$ varies in a neighborhood around
$\theta_0$. Proportional, in this case, is sufficient. We only need to compute the
\textit{relative} magnitudes of the task gradients in order to approximate Equation
\ref{clw}, since the loss weights $w_i$ are linearly scaled to fit Equation
\ref{weight_sum_constraint}.

To implement the approximation described by Theorem \ref{slaw_thm}, we simply keep a
running estimate of the standard deviation of $L_i$ through consecutive training steps.
Treating each $L_i$ as a random variable, we compute an exponential moving average of
the first and second moments of $L_i$, and use this to approximate the standard
deviation of $L_i$:
\begin{align}
    a_i &\gets \beta a_i + (1 - \beta) L_i^2, \label{slaw_estimate_1} \\
    b_i &\gets \beta b_i + (1 - \beta) L_i, \label{slaw_estimate_2} \\
    s_i &\gets \sqrt{a_i - b_i^2}, \label{slaw_estimate_3}
\end{align}
where $\beta$ is a parameter of the moving average. Since $s_i$ estimates the standard
deviation of $L_i$, by Theorem \ref{slaw_thm}, $s_i$ is nearly proportional to
$\|g_i\|$. We can then use $s_i$ as an approximation to $\|g_i\|$ in Equation \ref{clw},
which gives us SLAW:
\begin{align} \label{slaw}
    w_i = \frac{n}{s_i} \bigg/ \sum_{j=1}^n \frac{1}{s_j}
\end{align}
In the full algorithm, shown in Algorithm \ref{slaw_alg}, we replace $s_i$ with
$\max(s_i, 10^{-5})$ for numerical stability. The only hyperparameter of SLAW is the
moving average coefficient $\beta$.

\begin{algorithm}[t]
    \caption{Scaled Loss Approximate Weighting}
    \begin{algorithmic} \label{slaw_alg}
        \STATE $\theta \gets$ initialize()
        \STATE $w_i \gets 1.0$ for $i = 1$ to $n$
        \STATE $a_i, b_i, s_i \gets 0.0$ for $i = 1$ to $n$
        \FOR {$t = 0, 1, 2, ...$}
            \STATE $x \gets$ sample\_batch()
            \STATE $L_1, L_2, ..., L_n \gets$ forward($x$)
            \STATE $a_i \gets \beta a_i + (1 - \beta) L_i$ for $i = 1$ to $n$
            \STATE $b_i \gets \beta b_i + (1 - \beta) L_i^2$ for $i = 1$ to $n$
            \STATE $s_i \gets \max(\sqrt{a_i - b_i^2}, 10^{-5})$ for $i = 1$ to $n$
            \STATE $w_i \gets \frac{n}{s_i} / \sum_{j=1}^n \frac{1}{s_j}$ for $i = 1$ to $n$
            \STATE $L \gets \sum_{i=1}^n w_i L_i$
            \STATE $\theta \gets$ update($\theta, L$)
        \ENDFOR
    \end{algorithmic}
\end{algorithm}

Lastly, it should be noted that the approximation of $\|g_i\|$ by $s_i$ in SLAW does not
exactly match the conditions specified by Theorem \ref{slaw_thm}. Theorem \ref{slaw_thm}
says that we can approximate the norm of a function's gradient at a given point by
measuring the standard deviation of that function \textit{in a neighborhood} around that
point. In SLAW, we approximate the gradients of task losses by measuring the standard
deviation of each task loss not in a neighborhood around the current $\theta$, but over
recent values of $\theta$ during training. However, an empirical comparison shows that
SLAW's estimates of task gradient norms are still reasonably proportional to the true
gradient norm during training (see Appendix \ref{slaw_validation_appendix} for details).

\subsection{Relation to Existing Methods} \label{relation}
SLAW has a strong relationship with GradNorm \cite{chen_2017}, Dynamic Weight Averaging
\cite{liu_2019a}, and Adam \cite{kingma_2014}. We discuss their similarities and
differences below.

\textbf{GradNorm} \cite{chen_2017} is a well-known method for loss weighting in MTL with
a similar motivation as SLAW: to balance the norms of the gradients of each task's loss
function. GradNorm learns loss weights $w_i$ by minimizing the distance from the
weighted gradient norms to target values. Specifically, for task gradients $g_i = \nabla
L_i$, let $G = \frac{1}{n} \sum_{i=1}^n \|g_i\|$, let $\tilde{L}_i = L_i / L_{0i}$,
where $L_{0i}$ is the loss of the $i$-th task on the first step of training, and let
$r_i = \tilde{L}_i \big/ \frac{1}{n} \sum_{j=1}^n \tilde{L}_j$. The GradNorm loss
weights $w_i$ are learned by minimizing the following loss function:
\begin{align}
    L_{\text{grad}}(\{w_i\}_{i=1}^n) = \sum_{i=1}^n | w_i \|g_i\| - G r_i^{\alpha} |
\end{align}
where $\alpha$ is a hyperparameter that controls the degree to which the magnitude of
gradients from different tasks may differ under GradNorm. After each optimization step,
the loss weights are scaled so that their average value is 1.

On the surface, the GradNorm algorithm doesn't bear much resemblance to SLAW. We can
gain insight into the relationship between these two algorithms by directly computing
the global minimum of $L_{\text{grad}}$. We know that $L_{\text{grad}} \geq 0$, since
$L_{\text{grad}}$ is a sum of absolute values. We can achieve $L_{\text{grad}} = 0$ with
the following values of $w_i$:
\begin{align} \label{gradnorm_solution}
    w_i = \frac{G r_i^{\alpha}}{\|g_i\|}
\end{align}
Further, these are the only loss weights with $L_{\text{grad}} = 0$. Therefore, Equation
\ref{gradnorm_solution} gives a unique minimizer of $L_{\text{grad}}$. Scaling these
weights so that their average value is 1, we see that GradNorm's optimal loss weights
are:
\begin{align} \label{normalized_gradnorm_solution}
    w_i &= n \frac{r_i^{\alpha}}{\|g_i\|} \bigg/ \sum_{j=1}^n \frac{r_j^{\alpha}}{\|g_j\|}
\end{align}
In the case where $\alpha = 0$, where weights are learned so that all tasks have equal
gradient norms, this becomes:
\begin{align} \label{normalized_gradnormzero_solution}
    w_i = \frac{n}{\|g_i\|} \bigg/ \sum_{j=1}^n \frac{1}{\|g_j\|}
\end{align}
In summary, Equation \ref{normalized_gradnormzero_solution} describes the global
minimizer of $L_{\text{grad}}$ in the case that $\alpha = 0$. However, Equation
\ref{normalized_gradnormzero_solution} is the exact same as Equation \ref{clw}, which is
approximated by SLAW. Therefore, SLAW approximates the global solution to the GradNorm
loss weight optimization problem for the case that $\alpha = 0$. SLAW does this without
computing the task-specific gradients $g_i$, greatly reducing computational cost.

\textbf{Dynamic Weight Averaging} \cite{liu_2019a} was introduced as a simpler and
faster alternative to GradNorm, similarly to SLAW. It works by assigning more weight to
tasks for which learning is slow. The learning speed of a task is defined as the ratio
of that task's loss function on recent training steps: $L_i(t - 1) / L_i(t
- 2)$. DWA is efficient: it does not require individual task gradients.

However, our experiments show that DWA does not learn meaningful loss weights. For a
variety of tasks and architectures, we found that each loss weight $w_i$ learned by DWA
oscillates around 1.0. DWA isn't theoretically justified, and the empirical results from
\cite{liu_2019a} do not show that DWA outperforms the naive constant weight method with
statistical significance. Our theoretical and empirical evidence shows that SLAW
provides what DWA does not: an efficient, effective alternative to GradNorm.

\textbf{Adam} \cite{kingma_2014} is a widely-used stochastic optimization method. Adam,
like SLAW, keeps estimates of gradient statistics that are used to normalize model
updates:
\begin{align}
    \theta \gets \theta - \eta \cdot \hat{m}_t / (\sqrt{\hat{v}_t} + \epsilon)
\end{align}
where $\eta$ is the learning rate,  $\epsilon$ is a small constant included for
numerical stability, and $\hat{m}_t$ and $\hat{v}_t$ are running estimates of the first
and second moments, respectively, of the gradient of the function being optimized.
Adam's scaling of the update by a factor of $1 / (\sqrt{\hat{v}_t} + \epsilon)$ is very
similar to SLAW's, which approximately scales the update from each loss function $L_i$
by a factor of $1 / \|g_i\|$ (before normalization of $w_i$).

The key difference between Adam and SLAW is the fact that SLAW scales task loss
functions, whereas Adam scales updates for each model parameter. These two
normalizations have independent effects: SLAW balances training of the entire model
\textit{across tasks}, while Adam balances training on the aggregate MTL task
\textit{across parameters}.

\section{Experiments} \label{experiments}
With our experiments, we seek to answer the following questions: (1) How accurately and
quickly do loss weights learned by SLAW converge to the ideal loss weights compared to
existing methods? (2) How does the performance of multi-task networks trained with SLAW
compare to that of existing methods? (3) How does SLAW's running time compare to that of
existing methods?

We investigate each one of these questions by evaluating SLAW and various baselines on a
different application domain. To evaluate loss weight quality, we use a set of
non-linear regression tasks proposed in \cite{chen_2017}, where the ideal loss weights
are known. We refer to this collection of tasks as MTRegression. For overall
performance, we train jointly on semantic segmentation, surface normal estimation, and
depth prediction on the NYUv2 dataset \cite{silberman_2012}. Finally, to evaluate the
efficiency of SLAW relative to other methods, we train models on a number of virtual
screening tasks varying from 32 to 128 on the PubChem BioAssay (PCBA) dataset
\cite{wang_2016}.

In these investigations, we compare SLAW against the following multi-task baselines:
\begin{itemize}
    \item Constant: The naive loss weighting method, where $w_i = 1$ for all $i$.
    \item IdealConstant: Only used for MTRegression. Constant weight values of $w_i = 1
        / \sigma_i^2$, so that each loss function is perfectly balanced. This is used as
        an upper bound on the performance of the other baselines.
    \item Weighting by Uncertainty \cite{kendall_2017}: The loss weights $w_i$ are
        treated as learnable parameters which are trained jointly with model parameters
        $\theta$. The loss function is derived from a probabilistic formulation of
        classification and regression problems.
    \item GradNorm \cite{chen_2017}: Loss weights are learned by solving an
        optimization problem to balance the norm of task gradients. Details are given in
        Section \ref{relation}.
    \item Dynamic Weight Averaging \cite{liu_2019a}: DWA sets loss weights $w_i$ so that
        tasks for which learning has been slow will receive more weight.
    \item PCGrad \cite{yu_2020a}: PCGrad computes the gradients for each task loss and
        modifies them to avoid pairwise conflicts without explicitly computing loss
        weights.
\end{itemize}

Experiments were run in a unified codebase containing our own implementations of
Weighting by Uncertainty, GradNorm, DWA, PCGrad, and SLAW. Our GradNorm implementation
computes task-specific gradients only for the last shared layer of the network,
following the original work \cite{chen_2017}. The unified code ensures a fair comparison
between methods, and the code is publicly
available\footnote{\url{https://github.com/mtcrawshaw/meta}} for transparency and
reproducibility. Experiments were run on a single NVIDIA RTX 3090.

\subsection{Loss Weight Quality (MTRegression)}

\subsubsection{Training Settings}
\cite{chen_2017} introduces a collection of $n$ synthetic non-linear regression tasks
defined by the following functions:
\begin{align}
    f_i(\mathbf{x}) = \sigma_i \tanh((B + \epsilon_i) \mathbf{x})
\end{align}
where the matrix $B$ is shared between all tasks, while $\sigma_i$ and $\epsilon_i$ are
specific to each task. We refer to this collection of tasks as MTRegression. The task is
defined so that loss functions of various tasks have significantly different scale,
posing a testbed for multi-task optimization. The input and output dimensions are 250
and 100, respectively, and the elements of $B$ and $\epsilon_i$ are sampled
independently from $\mathcal{N}(0, 10)$ and $\mathcal{N}(0, 3.5)$, respectively. In our
experiments, we set $\sigma_i = i$ and $n = 10$. The dataset has 9000 training samples
and 1000 testing samples, where each sample has an input value and a label for each of
the $n$ tasks.

Following \cite{chen_2017}, we train a fully connected network made of a shared trunk
with 4 fully-connected layers, ReLU activations, and a hidden size of 100, followed by
$n$ one-layer task-specific output heads that produce the prediction for each task. Each
task's loss function is a standard squared error between model output and ground-truth.
In each experiment the network is trained to minimize the weighted sum of the $n$ task
losses: $\sum_{i=1}^n w_i L_i$, where $w_i$ is determined differently by each method
evaluated.

We run training for 300 epochs with a batch size of 304, following the table of
hyperparameters in Appendix \ref{hyperparameters}. For each method, we run 10 training
runs with different random seeds and report the average metric values over all runs.

We evaluate both the model performance and the quality of learned loss weights for each
method considered. Since the task loss functions $L_i$ are artificially scaled by a
factor of $\sigma_i^2$, the ideal value of each $w_i$ to balance the magnitude of task
loss functions is $w_i = 1 / \sigma_i^2$. Using this information, we define a loss
weight error of a set of loss weights $\{w_i\}_{i=1}^n$ as the mean squared error from
$\{w_i\}_{i=1}^n$ to the ideal weights. To evaluate performance on MTRegression, we
follow \cite{chen_2017} and compute a normalized loss that considers all tasks equally:
the mean of loss functions $L_i$ scaled by a factor of $1 / \sigma_i^2$.

\begin{table}[t]
\begin{center}
\begin{tabular}{|c||cc|}
\hline
 & Train NL & Test NL \\
\hline\hline
Constant & $13.036 \pm 0.032$ & $13.635 \pm 0.035$ \\
IdealConstant & $\mathbf{11.744} \pm 0.033$ & $\mathbf{12.274} \pm 0.035$ \\
Uncertainty & - & - \\
GradNorm & $11.955 \pm 0.039$ & $12.509 \pm 0.041$ \\
DWA & $13.036 \pm 0.032$ & $13.635 \pm 0.035$ \\
PCGrad & $12.986 \pm 0.037$ & $13.581 \pm 0.042$ \\
\hline
SLAW (Ours) & $\mathbf{11.762} \pm 0.034$ & $\mathbf{12.294} \pm 0.046$ \\
\hline
\end{tabular}
\end{center}
\caption{
    Final normalized loss (NL) for various methods on MTRegression, averaged over 10
    training runs and with 95\% confidence intervals. Best results for training and
    testing are bolded. Uncertainty \cite{kendall_2017} exhibited training instability.
}
\label{quality_table}
\end{table}

\subsubsection{Results}
\begin{figure}[t]
\begin{center}
   \includegraphics[width=0.8\linewidth]{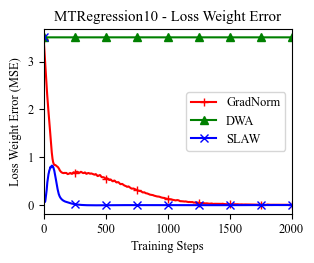}
\end{center}
\caption{
    Loss weight error on MTRegression, or the MSE between current loss weights and ideal
    weights, averaged over 10 training runs. For visual clarity, the loss weight error
    is shown for only the first 20\% of training.
}
\label{quality_figure}
\end{figure}

Figure \ref{quality_figure} shows the loss weight error of each method throughout
training, averaged over 10 training runs, and the normalized losses achieved at the end
of training are shown in Table \ref{quality_table}. The loss weight error is only
plotted for the first 2000 of 9900 training steps in order to better visualize the
differences between methods, since the loss weight errors remained stable after this
point in training. Since PCGrad doesn't explicitly compute loss weights, we do not
include it in the comparison of loss weight error.

SLAW outperforms all non-oracle baselines in terms of loss weight quality. Both SLAW and
GradNorm are able to learn ideal loss weights, but SLAW's weights converge faster than
GradNorm. This is likely because GradNorm iteratively tunes the weights to minimize
$L_{\text{grad}}$ while SLAW directly estimates a solution to the minimization.  On the
other hand, DWA is unable to learn meaningful loss weights, and its loss weight error
does not decrease throughout training.

Further, SLAW achieves the lowest normalized loss among all non-oracle baselines, and is
even competitive with the oracle IdealConstant. We conclude that SLAW can ideally
balance learning between tasks, and achieves this balance during training faster than
existing methods.

Results for Weighting by Uncertainty \cite{kendall_2017} are omitted, because training
with this method became unstable after the loss became sufficiently small. As pointed
out in \cite{chen_2017}, Weighting by Uncertainty yields values of $w_i$ proportional to
$1 / L_i$. If the loss gets very low, then the loss weights will grow similarly large,
resulting in a huge effective learning rate and therefore instability.  We believe that
this issue will not arise in real-world problems, where the loss will not become
extremely small.

\subsection{Performance (NYUv2)} \label{NYUv2_details}
\begin{table*}[t]
\begin{center}
\begin{tabular}{|c||cc|ccc|c|}
\hline
 & Avg Acc & Stdev Acc & Seg Pixel Acc & SN Acc & Depth Acc & Step Time \\
\hline\hline
Constant & $61.469 \pm 0.253$ & $9.542$ & $\mathbf{69.510} \pm 0.164$ & $48.280 \pm 0.452$ & $\mathbf{66.962} \pm 0.308$ & $\mathbf{0.573}$ \\
Uncertainty & $61.074 \pm 0.308$ & $9.602$ & $\mathbf{69.277} \pm 0.321$ & $47.950 \pm 0.406$ & $66.467 \pm 0.414$ & $0.577$ \\
GradNorm & $\mathbf{62.286} \pm 0.329$ & $8.612$ & $69.167 \pm 0.126$ & $\mathbf{50.547} \pm 0.552$ & $\mathbf{67.646} \pm 0.448$ & $1.243$ \\
DWA & $61.219 \pm 0.297$ & $9.656$ & $\mathbf{69.529} \pm 0.169$ & $48.071 \pm 0.403$ & $66.705 \pm 0.336$ & $0.587$ \\
PCGrad & $61.313 \pm 0.242$ & $9.349$ & $\mathbf{69.699} \pm 0.178$ & $48.372 \pm 0.233$ & $66.686 \pm 0.357$ & $1.133$ \\
\hline
SLAW (Ours) & $\mathbf{61.798} \pm 0.363$ & $\mathbf{7.917}$ & $69.133 \pm 0.124$ & $\mathbf{51.238} \pm 0.417$ & $65.623 \pm 0.670$ & $0.592$ \\
\hline
\end{tabular}
\end{center}
\caption{
    Multi-task NYUv2 test results for semantic segmentation, surface normal estimation,
    and depth prediction, averaged over 10 runs with 95\% confidence intervals. The
    result of the best performing method is bolded, as well any result whose confidence
    interval intersects that of the best result. GradNorm and SLAW achieve the highest
    average accuracy, though each training step with GradNorm takes over twice as long
    as any other method. SLAW exhibits the most equity across the 3 tasks.
}
\label{performance_table}
\end{table*}

\subsubsection{Training Settings}
NYUv2 \cite{silberman_2012} consists of 1449 RGB images which are densely labeled for
semantic segmentation (13 classes), surface normal estimation, and depth prediction. The
795 training images and 654 testing images depict indoor scenes with resolution 640x480,
though we downscale the images to 160x120.

Our multi-task architecture for NYUv2 is designed following \cite{kendall_2017}, with
some simplifications. The network consists of a feature extractor shared by all tasks,
followed by task-specific output branches. The shared feature extractor is a ResNet-50
\cite{he_2016} pretrained on ImageNet classification \cite{russakovsky_2015}, cut off
before the final average pooling layer. Each task-specific branch has 3 3x3 Conv-BN-ReLU
layers with 256 channels in the first two layers.

This unified architecture is trained by minimizing the weighted sum of task-specific
losses: $w_1 L_1 + w_2 L_2 + w_3 L_3$. Here $L_1, L_2$, and $L_3$ represent the loss
functions for semantic segmentation, surface normal estimation, and depth prediction
respectively. $L_1$ is the pixel-wise mean of the cross entropy loss, $L_2$ is the
negative pixel-wise mean of cosine similarity, and $L_3$ is the scale-invariant loss for
depth prediction used for training in \cite{eigen_2014}.

We train for 20 epochs with a batch size of 36, following the table of hyperparameters
given in Appendix \ref{hyperparameters}. For each method, we run 10 training runs with
different random seeds and report the average metric values over the 10 runs.

To evaluate all tasks on common ground, we define an accuracy metric for each task. For
semantic segmentation we measure the percentage of correctly classified pixels.
Performance on normal estimation is measured as the percentage of pixels for which the
angle between the ground truth normal and the predicted normal is less than 11.25
degrees. Lastly, for depth prediction we define accuracy as the percentage of pixels for
which the ratio between the ground truth and predicted depth is between 0.8 and 1.25. We
then report the mean of task accuracies as a measure of overall performance and the
standard deviation of task accuracies as a measure of inequity across tasks.

\subsubsection{Results}
Results are shown in Table \ref{performance_table}. SLAW and GradNorm achieve the
highest average accuracy across the three tasks, with a statistically significant
difference between these two methods and the other baselines. In addition, the spread of
accuracies across the three tasks is smallest when training with SLAW. Therefore, SLAW
exhibits the best overall performance and the most equitable training of tasks.

However, the increased average performance of SLAW and GradNorm doesn't come from an
increase on every task; these two methods incur a trade-off between task performances.
For example, SLAW and GradNorm are superior in surface normal estimation and depth
prediction, respectively, but both methods are outperformed by the other baselines in
semantic segmentation.

While GradNorm reaches the highest average accuracy, each of its training steps requires
more than double the time of any other method. SLAW retains the strong performance of
GradNorm while enjoying a significant boost in speed.

\subsection{Efficiency (PCBA)}
\begin{figure*}[t]
\begin{center}
   \includegraphics[width=0.35\linewidth]{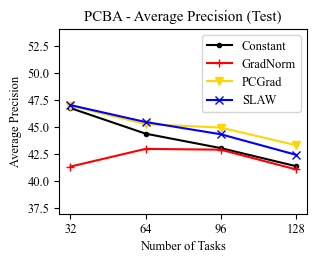}
   \includegraphics[width=0.34\linewidth]{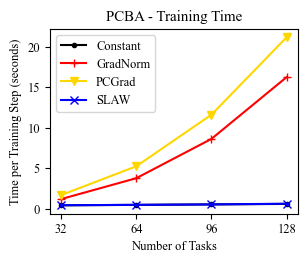}
\end{center}
\caption{
    Average precision (AP) and time per training step vs. number of tasks on PCBA. As
    the number of tasks grows from 32 to 128, training time stays nearly constant for
    SLAW, while the train step times for GradNorm and PCGrad grow by factors of 13.1 and
    12.2, respectively (right). SLAW achieves a higher AP than GradNorm and is
    competitive with PCGrad (left).
}
\label{efficiency_figure}
\end{figure*}

\subsubsection{Training Settings}
The PubChem BioAssay (PCBA) dataset \cite{wang_2016} contains data for 128 virtual
screening tasks. Virtual screening \cite{rollinger_2008} is the process of predicting
whether a candidate molecule will bind to a biological target in order to identify
promising compounds for new drugs. The 128 tasks are binary classification of candidate
molecules as either active or inactive for 128 different biological targets.  In total,
about 440K candidate molecules are labeled, though each molecule is labeled for an
average of 78 tasks, yielding over 34 million labeled candidate/target pairs. The
classes are not balanced; on average, only about 2\% of input molecules will bind to a
given target. We use the PCBA data hosted in the DeepChem \cite{ramsundar_2019}
repository.

Each input molecule is featurized as a 2048-length binary vector using RDKit
\cite{landrum} to compute ECFP features \cite{rogers_2010} (radius 4). We use the
Pyramidal architecture of \cite{ramsundar_2015}: a fully-connected network made of a
2-layer shared feature extractor and 1-layer task-specific output heads. The feature
extractor layers have 2000 and 100 units, respectively. Our loss function is the
weighted sum of the cross-entropy classification loss for each task. To account for
class imbalance, we scale the loss for each class by a factor inversely proportional to
the number of samples of that class.

We train on datasets with varying number of tasks and analyze each method's computation
time as a function of the number of tasks. From the 128 tasks in PCBA, we construct four
multi-task datasets with 32, 64, 96, and 128 tasks. Since some molecules aren't labeled
for any of the tasks in the first three datasets, we restrict these experiments to the
subset of input molecules which are labeled for at least one of the first 32 tasks,
leaving about 390K input molecules. Of these remaining inputs, we use 90\% for training
and 10\% for testing. For each dataset, we train for 100 epochs with a batch size of
40000, following the table of hyperparameters from Appendix \ref{hyperparameters}.
Efficiency and performance are evaluated with time per training step and the mean of the
Average Precision (AP) over all tasks, averaged over 5 random seeds.

\subsubsection{Results}
Results are shown in Figure \ref{efficiency_figure}. The training step time for GradNorm
and PCGrad grows significantly with the number of tasks, since these methods compute one
(at least partial) backwards pass per task for each training step. As the number of
tasks grows from 32 to 128, the train step times for GradNorm and PCGrad each grow by a
factor of more than 12, taking 16.2 seconds and 21.1 seconds, respectively, per training
step on 128 tasks. In contrast, the train step time of SLAW is nearly identical to
Constant, increasing only by a factor of 1.4 from 32 to 128 tasks. When training on the
full task set, SLAW trains 24.4 times faster than GradNorm and 33.0 times faster than
PCGrad.

Despite the efficiency gains, SLAW consistently achieves a larger average precision than
GradNorm, and is competitive with PCGrad. Interestingly, GradNorm is outperformed by
even the naive baseline Constant, but the gap between these methods shrinks as the
number of tasks increases. We conclude that SLAW can be efficiently scaled to a large
number of tasks without sacrificing performance.

\section{Conclusion} \label{conclusion}
In this paper, we introduced Scaled Loss Approximate Weighting (SLAW), an efficient and
general method for multi-task optimization that requires little to no hyperparameter
tuning. SLAW matches the performance of existing methods, but runs much faster by
estimating the norms of the gradients of each task's loss function instead of computing
gradients directly. We provided theoretical and empirical justification for the
estimation of these gradient norms, and we showed that SLAW directly estimates a
solution to a special case of the optimization problem that GradNorm \cite{chen_2017}
iteratively solves. Experimental results on a collection of non-linear regression tasks,
NYUv2, and virtual screening for drug discovery show that SLAW can learn to ideally
balance learning between tasks, matches or exceeds the performance of existing methods,
is much more efficient than existing methods, and is scalable to many tasks.

Efficiency and scalability to large collections of tasks are key qualities for MTL
algorithms. Until MTL \cite{zhang_2017, crawshaw_2020} and the related fields of
meta-learning \cite{hospedales_2021} and lifelong learning \cite{delange_2021} reach
their potential, we will be trapped in the era of gathering huge labeled datasets and
slowly retraining from scratch on each new task. The compute and data efficiency gains
afforded by MTL may help us leave this era, but making this a reality will require
improving the speed, performance, and flexibility of MTL methods.

\section*{Acknowledgements}
We would like to thank Gregory J. Stein, Abhishek Paudel, Raihan Islam Arnob, Roshan
Dhakal, and Sulabh Shrestha for feedback on an early draft, and Daniel Brogan for
feedback on the formatting of the proof of Theorem \ref{slaw_thm}.

\bibliographystyle{plain}
\bibliography{slaw}

\begin{thebibliography}{10}

\bibitem{achille_2019}
Alessandro Achille, Michael Lam, Rahul Tewari, Avinash Ravichandran, Subhransu
  Maji, Charless~C Fowlkes, Stefano Soatto, and Pietro Perona.
\newblock Task2vec: Task embedding for meta-learning.
\newblock In {\em Proceedings of the IEEE/CVF International Conference on
  Computer Vision}, pages 6430--6439, 2019.

\bibitem{alonso_2016}
H{\'e}ctor~Mart{\'\i}nez Alonso and Barbara Plank.
\newblock When is multitask learning effective? semantic sequence prediction
  under varying data conditions.
\newblock {\em arXiv preprint arXiv:1612.02251}, 2016.

\bibitem{caruana_1997}
Rich Caruana.
\newblock Multitask learning.
\newblock {\em Machine Learning}, 28(1):41--75, 1997.

\bibitem{chaudhry_2018}
Arslan Chaudhry, Marc'Aurelio Ranzato, Marcus Rohrbach, and Mohamed Elhoseiny.
\newblock Efficient lifelong learning with a-gem.
\newblock {\em arXiv preprint arXiv:1812.00420}, 2018.

\bibitem{chen_2017}
Zhao Chen, Vijay Badrinarayanan, Chen-Yu Lee, and Andrew Rabinovich.
\newblock Gradnorm: Gradient normalization for adaptive loss balancing in deep
  multitask networks, 2017.

\bibitem{ching_2018}
Travers Ching, Daniel~S. Himmelstein, Brett~K. Beaulieu-Jones, Alexandr~A.
  Kalinin, Brian~T. Do, Gregory~P. Way, Enrico Ferrero, Paul-Michael Agapow,
  Michael Zietz, Michael~M. Hoffman, Wei Xie, Gail~L. Rosen, Benjamin~J.
  Lengerich, Johnny Israeli, Jack Lanchantin, Stephen Woloszynek, Anne~E.
  Carpenter, Avanti Shrikumar, Jinbo Xu, Evan~M. Cofer, Christopher~A.
  Lavender, Srinivas~C. Turaga, Amr~M. Alexandari, Zhiyong Lu, David~J. Harris,
  Dave DeCaprio, Yanjun Qi, Anshul Kundaje, Yifan Peng, Laura~K. Wiley, Marwin
  H.~S. Segler, Simina~M. Boca, S.~Joshua Swamidass, Austin Huang, Anthony
  Gitter, and Casey~S. Greene.
\newblock Opportunities and obstacles for deep learning in biology and
  medicine.
\newblock {\em Journal of The Royal Society Interface}, 15(141):20170387, 2018.

\bibitem{collobert_2008}
Ronan Collobert and Jason Weston.
\newblock A unified architecture for natural language processing: Deep neural
  networks with multitask learning.
\newblock In {\em Proceedings of the 25th International Conference on Machine
  Learning}, ICML '08, page 160–167, New York, NY, USA, 2008. Association for
  Computing Machinery.

\bibitem{crawshaw_2020}
Michael Crawshaw.
\newblock Multi-task learning with deep neural networks: A survey, 2020.

\bibitem{dai_2015}
Jifeng Dai, Kaiming He, and Jian Sun.
\newblock Instance-aware semantic segmentation via multi-task network cascades,
  2015.

\bibitem{delange_2021}
Matthias Delange, Rahaf Aljundi, Marc Masana, Sarah Parisot, Xu~Jia, Ales
  Leonardis, Greg Slabaugh, and Tinne Tuytelaars.
\newblock A continual learning survey: Defying forgetting in classification
  tasks.
\newblock {\em IEEE Transactions on Pattern Analysis and Machine Intelligence},
  2021.

\bibitem{duong_2015}
Long Duong, Trevor Cohn, Steven Bird, and Paul Cook.
\newblock Low resource dependency parsing: Cross-lingual parameter sharing in a
  neural network parser.
\newblock In {\em Proceedings of the 53rd Annual Meeting of the Association for
  Computational Linguistics and the 7th International Joint Conference on
  Natural Language Processing (Volume 2: Short Papers)}, pages 845--850,
  Beijing, China, July 2015. Association for Computational Linguistics.

\bibitem{eigen_2015}
David Eigen and Rob Fergus.
\newblock Predicting depth, surface normals and semantic labels with a common
  multi-scale convolutional architecture, 2015.

\bibitem{eigen_2014}
David Eigen, Christian Puhrsch, and Rob Fergus.
\newblock Depth map prediction from a single image using a multi-scale deep
  network.
\newblock In {\em 28th Annual Conference on Neural Information Processing
  Systems 2014, NIPS 2014}, pages 2366--2374. Neural information processing
  systems foundation, 2014.

\bibitem{gao_2019}
Yuan Gao, Jiayi Ma, Mingbo Zhao, Wei Liu, and Alan~L Yuille.
\newblock Nddr-cnn: Layerwise feature fusing in multi-task cnns by neural
  discriminative dimensionality reduction.
\newblock In {\em Proceedings of the IEEE Conference on Computer Vision and
  Pattern Recognition}, pages 3205--3214, 2019.

\bibitem{guo_2018}
Michelle Guo, Albert Haque, De-An Huang, Serena Yeung, and Li~Fei-Fei.
\newblock Dynamic task prioritization for multitask learning.
\newblock In Vittorio Ferrari, Martial Hebert, Cristian Sminchisescu, and Yair
  Weiss, editors, {\em Computer Vision -- ECCV 2018}, pages 282--299, Cham,
  2018. Springer International Publishing.

\bibitem{he_2016}
Kaiming He, Xiangyu Zhang, Shaoqing Ren, and Jian Sun.
\newblock Deep residual learning for image recognition.
\newblock In {\em 2016 IEEE Conference on Computer Vision and Pattern
  Recognition (CVPR)}, pages 770--778, 2016.

\bibitem{hospedales_2021}
Timothy~M Hospedales, Antreas Antoniou, Paul Micaelli, and Amos~J Storkey.
\newblock Meta-learning in neural networks: A survey.
\newblock {\em IEEE Transactions on Pattern Analysis and Machine Intelligence},
  2021.

\bibitem{kendall_2017}
Alex Kendall, Yarin Gal, and Roberto Cipolla.
\newblock Multi-task learning using uncertainty to weigh losses for scene
  geometry and semantics, 2017.

\bibitem{kingma_2014}
Diederik Kingma and Jimmy Ba.
\newblock Adam: A method for stochastic optimization.
\newblock {\em International Conference on Learning Representations}, 12 2014.

\bibitem{landrum}
Greg Landrum.
\newblock Rdkit: Open-source cheminformatics, 2006.

\bibitem{liu_2020}
Liyang Liu, Yi~Li, Zhanghui Kuang, Jing-Hao Xue, Yimin Chen, Wenming Yang,
  Qingmin Liao, and Wayne Zhang.
\newblock Towards impartial multi-task learning.
\newblock In {\em International Conference on Learning Representations}, 2020.

\bibitem{liu_2019a}
S.~{Liu}, E.~{Johns}, and A.~J. {Davison}.
\newblock End-to-end multi-task learning with attention.
\newblock In {\em 2019 IEEE/CVF Conference on Computer Vision and Pattern
  Recognition (CVPR)}, pages 1871--1880, 2019.

\bibitem{liu_2019b}
Shengchao Liu, Yingyu Liang, and Anthony Gitter.
\newblock Loss-balanced task weighting to reduce negative transfer in
  multi-task learning.
\newblock In {\em Proceedings of the AAAI Conference on Artificial
  Intelligence}, volume~33, pages 9977--9978, 2019.

\bibitem{lu_2017}
Yongxi Lu, Abhishek Kumar, Shuangfei Zhai, Yu~Cheng, Tara Javidi, and Rogerio
  Feris.
\newblock Fully-adaptive feature sharing in multi-task networks with
  applications in person attribute classification.
\newblock In {\em Proceedings of the IEEE conference on computer vision and
  pattern recognition}, pages 5334--5343, 2017.

\bibitem{miotto_2017}
Riccardo Miotto, Fei Wang, Shuang Wang, Xiaoqian Jiang, and Joel~T Dudley.
\newblock {Deep learning for healthcare: review, opportunities and challenges}.
\newblock {\em Briefings in Bioinformatics}, 19(6):1236--1246, 05 2017.

\bibitem{misra_2016}
Ishan Misra, Abhinav Shrivastava, Abhinav Gupta, and Martial Hebert.
\newblock Cross-stitch networks for multi-task learning.
\newblock In {\em Proceedings of the IEEE Conference on Computer Vision and
  Pattern Recognition}, pages 3994--4003, 2016.

\bibitem{silberman_2012}
Pushmeet~Kohli Nathan~Silberman, Derek~Hoiem and Rob Fergus.
\newblock Indoor segmentation and support inference from rgbd images.
\newblock In {\em ECCV}, 2012.

\bibitem{parisotto_2015}
Emilio Parisotto, Jimmy~Lei Ba, and Ruslan Salakhutdinov.
\newblock Actor-mimic: Deep multitask and transfer reinforcement learning.
\newblock {\em arXiv preprint arXiv:1511.06342}, 2015.

\bibitem{pinto_2017}
Lerrel Pinto and Abhinav Gupta.
\newblock Learning to push by grasping: Using multiple tasks for effective
  learning.
\newblock In {\em 2017 IEEE International Conference on Robotics and Automation
  (ICRA)}, pages 2161--2168. IEEE, 2017.

\bibitem{ramsundar_2019}
Bharath Ramsundar, Peter Eastman, Patrick Walters, Vijay Pande, Karl Leswing,
  and Zhenqin Wu.
\newblock {\em Deep Learning for the Life Sciences}.
\newblock O'Reilly Media, 2019.

\bibitem{ramsundar_2015}
Bharath Ramsundar, Steven Kearnes, Patrick Riley, Dale Webster, David
  Konerding, and Vijay Pande.
\newblock Massively multitask networks for drug discovery.
\newblock {\em arXiv preprint arXiv:1502.02072}, 2015.

\bibitem{rogers_2010}
David Rogers and Mathew Hahn.
\newblock Extended-connectivity fingerprints.
\newblock {\em Journal of Chemical Information and Modeling}, 50(5):742--754,
  April 2010.

\bibitem{rollinger_2008}
Judith~M. Rollinger, Hermann Stuppner, and Thierry Langer.
\newblock {\em Virtual screening for the discovery of bioactive natural
  products}, pages 211--249.
\newblock Birkh{\"a}user Basel, Basel, 2008.

\bibitem{ruder_2019}
Sebastian Ruder, Joachim Bingel, Isabelle Augenstein, and Anders S{\o}gaard.
\newblock Latent multi-task architecture learning.
\newblock In {\em Proceedings of the AAAI Conference on Artificial
  Intelligence}, volume~33, pages 4822--4829, 2019.

\bibitem{russakovsky_2015}
Olga Russakovsky, Jia Deng, Hao Su, Jonathan Krause, Sanjeev Satheesh, Sean Ma,
  Zhiheng Huang, Andrej Karpathy, Aditya Khosla, Michael Bernstein, et~al.
\newblock Imagenet large scale visual recognition challenge.
\newblock {\em International journal of computer vision}, 115(3):211--252,
  2015.

\bibitem{standley_2019}
Trevor Standley, Amir~R Zamir, Dawn Chen, Leonidas Guibas, Jitendra Malik, and
  Silvio Savarese.
\newblock Which tasks should be learned together in multi-task learning?
\newblock {\em arXiv preprint arXiv:1905.07553}, 2019.

\bibitem{wang_2016}
Yanli Wang, Stephen~H. Bryant, Tiejun Cheng, Jiyao Wang, Asta Gindulyte,
  Benjamin~A. Shoemaker, Paul~A. Thiessen, Siqian He, and Jian Zhang.
\newblock {PubChem BioAssay: 2017 update}.
\newblock {\em Nucleic Acids Research}, 45(D1):D955--D963, 11 2016.

\bibitem{yu_2020a}
Tianhe Yu, Saurabh Kumar, Abhishek Gupta, Sergey Levine, Karol Hausman, and
  Chelsea Finn.
\newblock Gradient surgery for multi-task learning.
\newblock {\em Advances in Neural Information Processing Systems}, 33, 2020.

\bibitem{zamir_2018}
Amir~R Zamir, Alexander Sax, William Shen, Leonidas~J Guibas, Jitendra Malik,
  and Silvio Savarese.
\newblock Taskonomy: Disentangling task transfer learning.
\newblock In {\em Proceedings of the IEEE conference on computer vision and
  pattern recognition}, pages 3712--3722, 2018.

\bibitem{zhang_2017}
Yu~Zhang and Qiang Yang.
\newblock A survey on multi-task learning.
\newblock {\em arXiv preprint arXiv:1707.08114}, 2017.

\bibitem{zhang_2014}
Zhanpeng Zhang, Ping Luo, Chen~Change Loy, and Xiaoou Tang.
\newblock Facial landmark detection by deep multi-task learning.
\newblock In David Fleet, Tomas Pajdla, Bernt Schiele, and Tinne Tuytelaars,
  editors, {\em Computer Vision -- ECCV 2014}, pages 94--108, Cham, 2014.
  Springer International Publishing.

\bibitem{zheng_2018}
Feng Zheng, Cheng Deng, Xing Sun, Xinyang Jiang, Xiaowei Guo, Zongqiao Yu,
  Feiyue Huang, and Rongrong Ji.
\newblock Pyramidal person re-identification via multi-loss dynamic training,
  2018.

\end{thebibliography}

\onecolumn
\appendix
\section{Proof of Theorem \ref{slaw_thm}} \label{slaw_thm_proof}

\begin{proof}
For a given $\mathbf{x}_0 \in \mathbb{R}^d$ and $\epsilon > 0$, let $\epsilon_1 =
\sqrt{\epsilon}$ and $\epsilon_2 = \frac{\epsilon}{4K}$. Let $\delta_1$ be such that
$|f(\mathbf{x}) - f(\mathbf{x}_0)| < \epsilon_1$ whenever $|\mathbf{x} - \mathbf{x}_0| <
\delta_1$ and let $\delta_2$ be such that $\| \nabla f(\mathbf{x}) - \nabla
f(\mathbf{x}_0) \| < \epsilon_2$ whenever $|\mathbf{x} - \mathbf{x}_0| < \delta_2$. Let
$\delta = \min(\delta_1, \delta_2, 1)$. Let $\mathbf{X}$ be a random variable which is
uniformly distributed over $\{\mathbf{x}_0 + \mathbf{d} : \|\mathbf{d}\| < \delta \}$.
Then let $K_1 = \mathbb{E} \left[ \cos^2 \theta ~\| \mathbf{X} - \mathbf{x}_0 \|^2
\right]$ (where $\theta$ is the angle between $\mathbf{X} - \mathbf{x}_0$ and $(1, 0,
..., 0, 0)$), $K_3 = \mathbb{E} \left[ \| \mathbf{X} - \mathbf{x}_0 \|^2 \right]$, and
$K_2 = \epsilon_2^2 K_3 - \epsilon_1^2/2$.

Now, let $\mu = \mathbb{E}[f(\mathbf{X})]$. Then by linearity of expectation:
\begin{align}
    \text{Var}[f(\mathbf{X})] &= \mathbb{E}[(f(\mathbf{X}) - \mu)^2] \\
    &= \mathbb{E}[((f(\mathbf{X}) - f(\mathbf{x}_0)) - (\mu - f(\mathbf{x}_0)))^2] \\
    &= \mathbb{E}[(f(\mathbf{X}) - f(\mathbf{x}_0))^2] - 2\mathbb{E}[(f(\mathbf{X}) - f(\mathbf{x}_0))(\mu - f(\mathbf{x}_0))] + (\mu - f(\mathbf{x}_0))^2 \\
    &= \mathbb{E}[(f(\mathbf{X}) - f(\mathbf{x}_0))^2] - (\mu - f(\mathbf{x}_0))^2, \label{thm_slaw_eq1}
\end{align}

We will bound $\text{Var}[f(\mathbf{\mathbf{X}})]$ by bounding each of the two terms in
Equation \ref{thm_slaw_eq1} individually. Starting with $(\mu - f(\mathbf{x}_0))^2$, we
know that $\mu = f(\mathbf{a})$ for some $\mathbf{a}$ with $\|\mathbf{a} - \mathbf{x}_0\| <
\delta$ by applying the intermediate value theorem. Therefore
\begin{align}
    (\mu - f(\mathbf{x}_0))^2 = (f(\mathbf{a}) - f(\mathbf{x}_0))^2 \leq \epsilon_1^2
\end{align}
where the upper bound comes from the fact that $\|\mathbf{a} - \mathbf{x}_0\| < \delta <
\delta_1$. This gives the bound
\begin{align} \label{thm_slaw_bound1}
    0 \leq (\mu - f(\mathbf{x}_0))^2 \leq \epsilon_1^2
\end{align}

To bound $\mathbb{E}[(f(\mathbf{X}) - f(\mathbf{x}_0))^2]$, the second term from
Equation \ref{thm_slaw_eq1}, let $\mathbf{x} \in \mathbb{R}^d$ with $\|\mathbf{x} -
\mathbf{x}_0\| < \delta$, $\mathbf{u} = (\mathbf{x} - \mathbf{x_0}) / \| \mathbf{x} -
\mathbf{x_0} \|$. For $t \in [0, \|\mathbf{x} - \mathbf{x}_0\|]$, let $\mathbf{x}_t =
\mathbf{x}_0 + t\mathbf{u}$ and $\mathbf{d}_t = \nabla f(\mathbf{x}_t) - \nabla
f(\mathbf{x}_0)$. By the fundamental theorem of calculus, we have:
\begin{align}
    |f(\mathbf{x}) - f(\mathbf{x}_0)| &= \left| \int_0^{\|\mathbf{x} - \mathbf{x}_0\|} D_{\mathbf{u}} f(\mathbf{x}_t) ~dt \right| \\
    &= \left| \int_0^{\|\mathbf{x} - \mathbf{x}_0\|} \nabla f(\mathbf{x}_t) \cdot \mathbf{u} ~dt \right| \\
    &= \left| \int_0^{\|\mathbf{x} - \mathbf{x}_0\|} \nabla f(\mathbf{x}_0) \cdot \mathbf{u} ~dt + \int_0^{\|\mathbf{x} - \mathbf{x}_0\|} \mathbf{d}_t \cdot \mathbf{u} ~dt \right| \\
    &= \left| \nabla f(\mathbf{x}_0) \cdot (\mathbf{x} - \mathbf{x}_0) + \int_0^{\|\mathbf{x} - \mathbf{x}_0\|} \mathbf{d}_t \cdot \mathbf{u} ~dt \right|, \label{thm_slaw_eq2}
\end{align}
Applying the triangle inequality to Equation \ref{thm_slaw_eq2}, we obtain:
\begin{align}
    |f(\mathbf{x}) - f(\mathbf{x}_0)| &\leq | \nabla f(\mathbf{x}_0) \cdot (\mathbf{x} - \mathbf{x}_0) | + \left| \int_0^{\|\mathbf{x} - \mathbf{x}_0\|} \mathbf{d}_t \cdot \mathbf{u} ~dt \right| \\
    &\leq | \nabla f(\mathbf{x}_0) \cdot (\mathbf{x} - \mathbf{x}_0) | + \int_0^{\|\mathbf{x} - \mathbf{x}_0\|} | \mathbf{d}_t \cdot \mathbf{u} | ~dt \\
    &\leq | \nabla f(\mathbf{x}_0) \cdot (\mathbf{x} - \mathbf{x}_0) | + \int_0^{\|\mathbf{x} - \mathbf{x}_0\|} \| \mathbf{d}_t \| ~dt \\
    &< | \nabla f(\mathbf{x}_0) \cdot (\mathbf{x} - \mathbf{x}_0) | + \epsilon_2 \|\mathbf{x} - \mathbf{x}_0\| \\
    &= | (\| \nabla f(\mathbf{x}_0) \| \cos \theta + \epsilon_2) \|\mathbf{x} - \mathbf{x}_0\|, \label{thm_slaw_bound2.1a}
\end{align}
where $\theta$ is the angle between $\nabla f(\mathbf{x}_0)$ and $\mathbf{x} -
\mathbf{x}_0$. We can obtain a similar lower bound by again applying the triangle
inequality in a slightly different form to Equation \ref{thm_slaw_eq2}:
\begin{align}
    |f(\mathbf{x}) - f(\mathbf{x}_0)| &\geq | \nabla f(\mathbf{x}_0) \cdot (\mathbf{x} - \mathbf{x}_0) | - \left| \int_0^{\|\mathbf{x} - \mathbf{x}_0\|} \mathbf{d}_t \cdot \mathbf{u} ~dt \right|
\end{align}
which, by an identical computation as above, leads to the lower bound:
\begin{align}
    |f(\mathbf{x}) - f(\mathbf{x}_0)| &> | (\| \nabla f(\mathbf{x}_0) \| \cos \theta - \epsilon_2) \|\mathbf{x} - \mathbf{x}_0\|, \label{thm_slaw_bound2.1b}
\end{align}
Therefore, from Equation \ref{thm_slaw_bound2.1a}, we have
\begin{align}
    \mathbb{E} \left[ (f(\mathbf{X}) - f(\mathbf{x}_0))^2 \right] &< \mathbb{E} \left[ (\| \nabla f(\mathbf{x}_0) \| \cos \theta - \epsilon_2)^2 \|\mathbf{X} - \mathbf{x}_0\|^2 \right] \\
    &< \| \nabla f(\mathbf{x}_0) \|^2 ~\mathbb{E} \left[ \cos^2 \theta ~\| \mathbf{X} - \mathbf{x}_0 \|^2 \right] +
      2 \epsilon_2 \| \nabla f(\mathbf{x}_0) \| ~\mathbb{E} \left[ \| \mathbf{X} - \mathbf{x}_0 \|^2 \right] + \epsilon_2^2 \mathbb{E} \left[ \| \mathbf{X} - \mathbf{x}_0 \|^2 \right]
\end{align}
Since $f$ is $K$-Lipschitz, we have $\| \nabla f(\mathbf{x}_0) \| \leq K$. So
\begin{align}
    \mathbb{E} \left[ (f(\mathbf{X}) - f(\mathbf{x}_0))^2 \right] < K_1 \| \nabla f(\mathbf{x}_0) \|^2 + 2 \epsilon_2 K K_3 + \epsilon_2^2 K_3, \label{thm_slaw_bound2a}
\end{align}
Performing the same operations with the lower bound of $|f(\mathbf{x}) -
f(\mathbf{x}_0)|$ from Equation \ref{thm_slaw_bound2.1b}, we obtain:
\begin{align}
    \mathbb{E} \left[ (f(\mathbf{X}) - f(\mathbf{x}_0))^2 \right] > K_1 \| \nabla f(\mathbf{x}_0) \|^2 - 2 \epsilon_2 K K_3 + \epsilon_2^2 K_3, \label{thm_slaw_bound2b}
\end{align}

We can now derive an upper bound for $\text{Var}[f(\mathbf{X})]$ by plugging Equations
\ref{thm_slaw_bound1} and \ref{thm_slaw_bound2a} into Equation \ref{thm_slaw_eq1}:
\begin{align}
    \text{Var}[f(\mathbf{X})] &< K_1 \| \nabla f(\mathbf{x}_0) \|^2 + 2 \epsilon_2 K K_3 + \epsilon_2^2 K_3 \\
    \text{Var}[f(\mathbf{X})] - ( K_1 \| \nabla f(\mathbf{x}_0) \|^2 + \epsilon_2^2 K_3 ) &< 2 \epsilon_2 K K_3 \\
    \text{Var}[f(\mathbf{X})] - ( K_1 \| \nabla f(\mathbf{x}_0) \|^2 + \epsilon_2^2 K_3  - \epsilon_1^2 / 2) &< 2 \epsilon_2 K K_3 + \epsilon_1^2 / 2, \label{thm_slaw_bound3a}
\end{align}
Similarly, to derive a lower bound for $\text{Var}[f(\mathbf{X})]$, we plug Equations
\ref{thm_slaw_bound1} and \ref{thm_slaw_bound2b} into Equation \ref{thm_slaw_eq1}:
\begin{align}
    \text{Var}[f(\mathbf{X})] &> K_1 \| \nabla f(\mathbf{x}_0) \|^2 - 2 \epsilon_2 K K_3 + \epsilon_2^2 K_3 - \epsilon_1^2 \\
    \text{Var}[f(\mathbf{X})] - ( K_1 \| \nabla f(\mathbf{x}_0) \|^2 + \epsilon_2^2 K_3) &> -2 \epsilon_2 K K_3 - \epsilon_1^2 \\
    \text{Var}[f(\mathbf{X})] - ( K_1 \| \nabla f(\mathbf{x}_0) \|^2 + \epsilon_2^2 K_3 - \epsilon_1^2 / 2) &> -2 \epsilon_2 K K_3 - \epsilon_1^2 / 2, \label{thm_slaw_bound3b}
\end{align}
Recall that $\delta < 1$, so $K_3 < 1$. Finally, combining Equations
\ref{thm_slaw_bound3a} and \ref{thm_slaw_bound3b}:
\begin{align}
    \left| \text{Var}[f(\mathbf{X})] - ( K_1 \| \nabla f(\mathbf{x}_0) \|^2 + \epsilon_2^2 K_3 - \epsilon_1^2 / 2) \right| &< 2 \epsilon_2 K K_3 + \epsilon_1^2 / 2 \\
    \left| \text{Var}[f(\mathbf{X})] - ( K_1 \| \nabla f(\mathbf{x}_0) \|^2 + \epsilon_2^2 K_3 - \epsilon_1^2 / 2) \right| &< 2 \epsilon_2 K + \epsilon_1^2 / 2 \\
    \left| \text{Var}[f(\mathbf{X})] - ( K_1 \| \nabla f(\mathbf{x}_0) \|^2 + K_2 ) \right| &< \epsilon / 2 + \epsilon / 2 \\
    \left| \text{Var}[f(\mathbf{X})] - ( K_1 \| \nabla f(\mathbf{x}_0) \|^2 + K_2 ) \right| &< \epsilon
\end{align}

\end{proof}

\section{Empirical Validation of SLAW} \label{slaw_validation_appendix}
We conduct an empirical validation of SLAW by checking the accuracy of SLAW's estimate
of gradient magnitudes. Specifically, for task gradient $g_i = \nabla L_i$, SLAW
estimates $\|g_i\|$ with $s_i$ as defined in Equations \ref{slaw_estimate_1},
\ref{slaw_estimate_2}, and \ref{slaw_estimate_3}. However, $s_i$ is not an unbiased
estimate of $\|g_i\|$. Theorem \ref{slaw_thm} only tells us that $s_i$ is nearly
proportional to $\|g_i\|$. For our case, nearly proportional is good enough. The end
goal of this estimation is not to estimate the magnitudes $\|g_i\|$ exactly, but to
estimate the relative magnitudes of each $\|g_i\|$ as described by Equation \ref{clw}.
We postulate that this can be estimated with the relative magnitudes of each $s_i$.

To evaluate whether SLAW can accurately estimate the weights specified by Equation
\ref{clw}, we collect samples of SLAW's estimation $s_i$ along with the true gradient
norms $\|g_i\|$ during training on the MTRegression dataset. For each sample of $s_i$
and $\|g_i\|$, we plot the pair $(x, y)$ defined as follows:
\begin{align}
    x &= \frac{n}{\|g_i\|} \bigg/ \sum_{j=1}^n \frac{1}{\|g_j\|} \\
    y &= \frac{n}{s_i} \bigg/ \sum_{j=1}^n \frac{1}{s_i}
\end{align}
Each sample is taken during a randomly chosen step of training between training step 10
and training step 1000. We cut off the sampling at training step 1000 in order to ensure
that the experiment ran in a reasonable amount of time. Also, each sample is generated
from a different training run, to ensure that all points on the plot are independently
generated. The color and marker of each data point corresponds to the task whose
gradient is being estimated by SLAW.

\begin{figure}[t]
\begin{center}
   \includegraphics[width=0.45\linewidth]{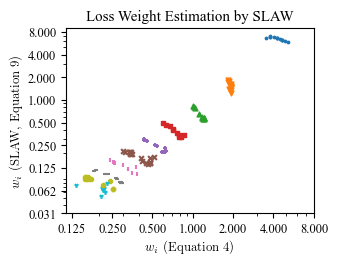}
\end{center}
\caption{
    Empirical validation of SLAW's estimate of the gradient norm while training on
    MTRegression dataset. Each data point shows the value of a loss weight $w_i$ as
    assigned by Equation \ref{clw} on a random training step (x-axis) and SLAW's weight
    value as assigned by Equation \ref{slaw} on the same step (y-axis). The color and
    marker shape of each point corresponds to the task index $i$ for which the true and
    estimated gradients are plotted. Each data point is generated from a different
    training run to ensure independence.
}
\label{slaw_validation_figure}
\end{figure}

Figure \ref{slaw_validation_figure} shows the result of plotting 120 such $(x, y)$
pairs. The figure shows that each $x$ provides a close approximation to the
corresponding $y$. From this, we conclude that SLAW computes a valid estimation of the
task gradient norms, and the loss weights learned by SLAW are close to the loss weights
specified by Equation \ref{clw}.

\section{Experiment Hyperparameters} \label{hyperparameters}
All hyperparameters specific to the loss weighting method were chosen to match the
values provided by the original papers.
\begin{table}[ht]
\begin{center}
\begin{tabular}{|c||ccc|}
\hline
 & MTRegression & NYUv2 & PCBA \\
\hline\hline
Epochs & 300 & 20 & 100 \\
Batch size & 304 & 36 & 40000 \\
Learning rate & 7e-4 & 7e-4 & 3e-4 \\
Gradient clipping (max gradient norm) & 0.5 & 0.5 & 0.5 \\
Backbone architecture & FC (4 layer) & ResNet-50 & Pyramidal \cite{ramsundar_2015} \\
Hidden size & 100 & N/A & 2000, 100 \\
Activation function & Tanh & ReLU & ReLU \\
Moving average coefficient (SLAW/DWA) & 0.99/0.9 & 0.99/0.9 & 0.99/0.9 \\
Loss weight LR (GradNorm) & 0.025 & 0.025 & 0.025 \\
Asymmetry (GradNorm) & 0.12 & 1.5 & 0.12 \\
Loss weight temperature (DWA) & 2.0 & 2.0 & 2.0 \\
\hline
\end{tabular}
\end{center}
\end{table}

\end{document}